\documentclass[conference]{IEEEtran}
\IEEEoverridecommandlockouts
\usepackage{cite}
\usepackage{amsmath,amssymb,amsfonts}
\usepackage{algorithmic}
\usepackage{graphicx}
\usepackage{textcomp}
\usepackage{xcolor}
\usepackage{multirow}

\usepackage{hyperref}

\usepackage[numbers]{natbib}

\def\BibTeX{{\rm B\kern-.05em{\sc i\kern-.025em b}\kern-.08em
    T\kern-.1667em\lower.7ex\hbox{E}\kern-.125emX}}
\begin{document}

\title{Spectral Enhancement and Pseudo-Anchor Guidance for Infrared-Visible Person Re-Identification}
\author{
Yiyuan Ge$^{1,\dag}$ \quad Zhihao Chen$^{1,\dag}$ \quad Ziyang Wang$^{2*}$ \quad Jiaju Kang$^{3}$ \quad Mingya Zhang$^{4}$\\
    $^{1}$ Beijing Information Science And Technology University \quad $^{2}$ University of Oxford \quad \\
    $^{3}$ Beijing Normal University \quad
    $^{4}$ Nanjing University \\
    1209027338@qq.com \quad 2021011561@bistu.edu.cn \quad ziyang.wang17@gmail.com \quad \\ kjj\_python@163.com \quad dg20330034@smail.nju.edu.cn
}

\maketitle

\begin{abstract}
The development of deep learning has facilitated the application of person re-identification (ReID) technology in intelligent security. Visible-infrared person re-identification (VI-ReID) aims to match pedestrians across infrared and visible modality images enabling 24-hour surveillance. Current studies relying on unsupervised modality transformations as well as inefficient embedding constraints to bridge the spectral differences between infrared and visible images, however, limit their potential performance. To tackle the limitations of the above approaches, this paper introduces a simple yet effective \textbf{S}pectral \textbf{E}nhancement and \textbf{P}seudo-anchor \textbf{G}uidance \textbf{Net}work, named \textbf{SEPG-Net}. Specifically, we propose a more homogeneous spectral enhancement scheme based on frequency domain information and greyscale space, which avoids the information loss typically caused by inefficient modality transformations. Further, a \textbf{P}seudo \textbf{A}nchor-guided \textbf{B}idirectional \textbf{A}ggregation \textbf{(PABA)} loss is introduced to bridge local modality discrepancies while better preserving discriminative identity embeddings. Experimental results on two public benchmark datasets demonstrate the superior performance of SEPG-Net against other state-of-the-art methods. The code is available at \textcolor{red}{https://github.com/1024AILab/ReID-SEPG}.
\end{abstract}


\section{Introduction}
\begin{figure}[t]
\begin{minipage}[b]{1.0\linewidth}
  \centering
  \centerline{\includegraphics[width=\columnwidth]{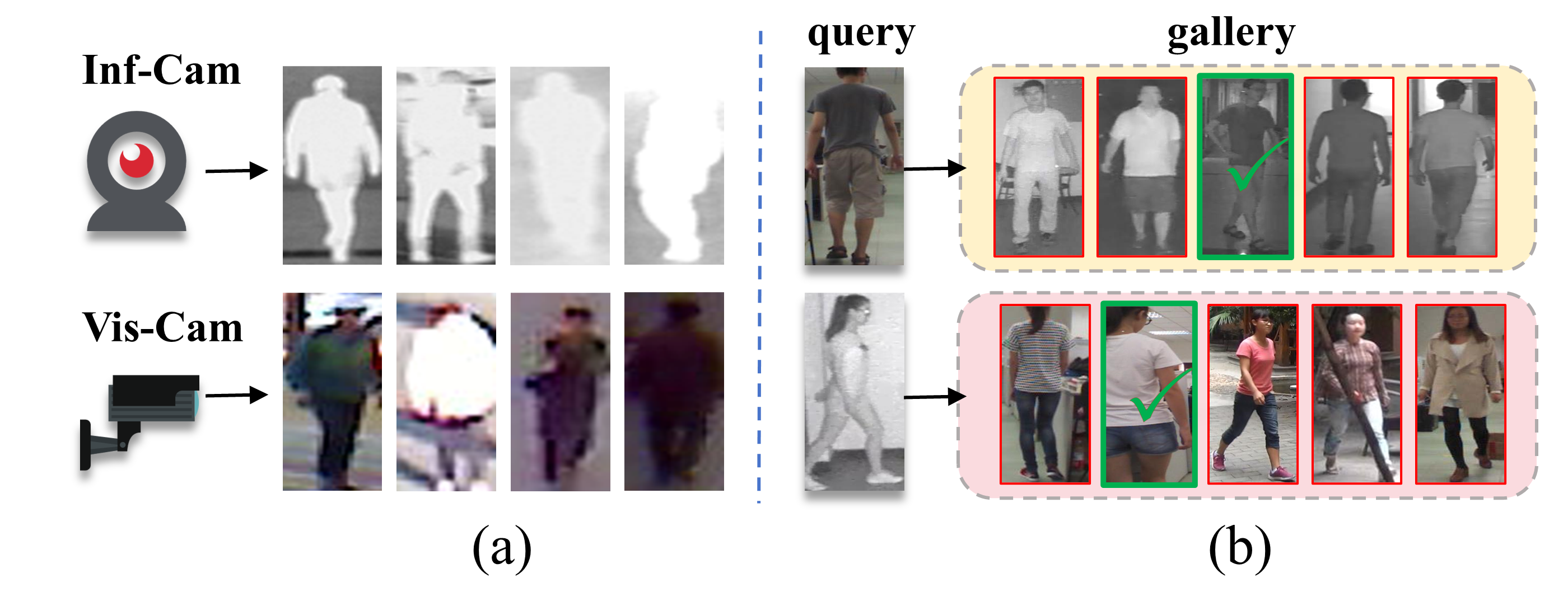}}
\end{minipage}
\caption{(a) Visible and infrared images are captured by visible cameras and dedicated infrared cameras, respectively (example images from the RegDB dataset\cite{b20}). (b) A brief illustration of the VI-ReID task (example images taken from the SYSU-MM01 dataset\cite{b21}).}
\label{fig:1}
\end{figure}

Person Re-identification (ReID) has attracted significant attention due to its wide applications in security and surveillance, aiming to recognize individuals across different cameras and scenes \cite{b1, b2, b3, b4}. Traditionally, ReID tasks are primarily performed under visible light conditions. Visible-Infrared Person Re-identification (VI-ReID) extends this task to nighttime scenarios by leveraging multi-modality cameras, making it essential for real-world applications. Pedestrian images captured by infrared and visible cameras differ significantly as illustrated in Fig. \ref{fig:1}(a). The infrared image highlights more distinct silhouette features, while the visible image provides richer color information. Fig. \ref{fig:1}(b) illustrates the goal of VI-ReID is to match a query target from the hetero-modality gallery. The large spectral gap between infrared and visible images, however, poses severe challenges for this task.

Following the above concern, several pioneering studies \cite{b5, b6, b7, b8} maintain a pre-trained Generative Adversarial Network (GAN) to achieve mutual transformation between infrared and visible images. However, this unsupervised generation is unstable, often compromising crucial modality information. Besides, other mainstream methods tackle the VI-ReID task by learning associative representations of infrared and visible images, using cross-centre losses to mitigate modality discrepancies \cite{b9, b10, b11, b12, b13}. Although these methods demonstrate effectiveness in some cases, certain limitations continue to hinder their performance. For example, \cite{b9, b10} only attempt to aggregate the centers of the two modalities, while \cite{b11} only focuses on intra-class center features and neglects modality categories. Moreover, these approaches tend to prioritize extracting modality-consistent representations while overlooking the distinctive identity embeddings.

\begin{figure*}[t!]
\begin{minipage}[b]{1.0\linewidth}
  \centering
  \centerline{\includegraphics[width=\columnwidth]{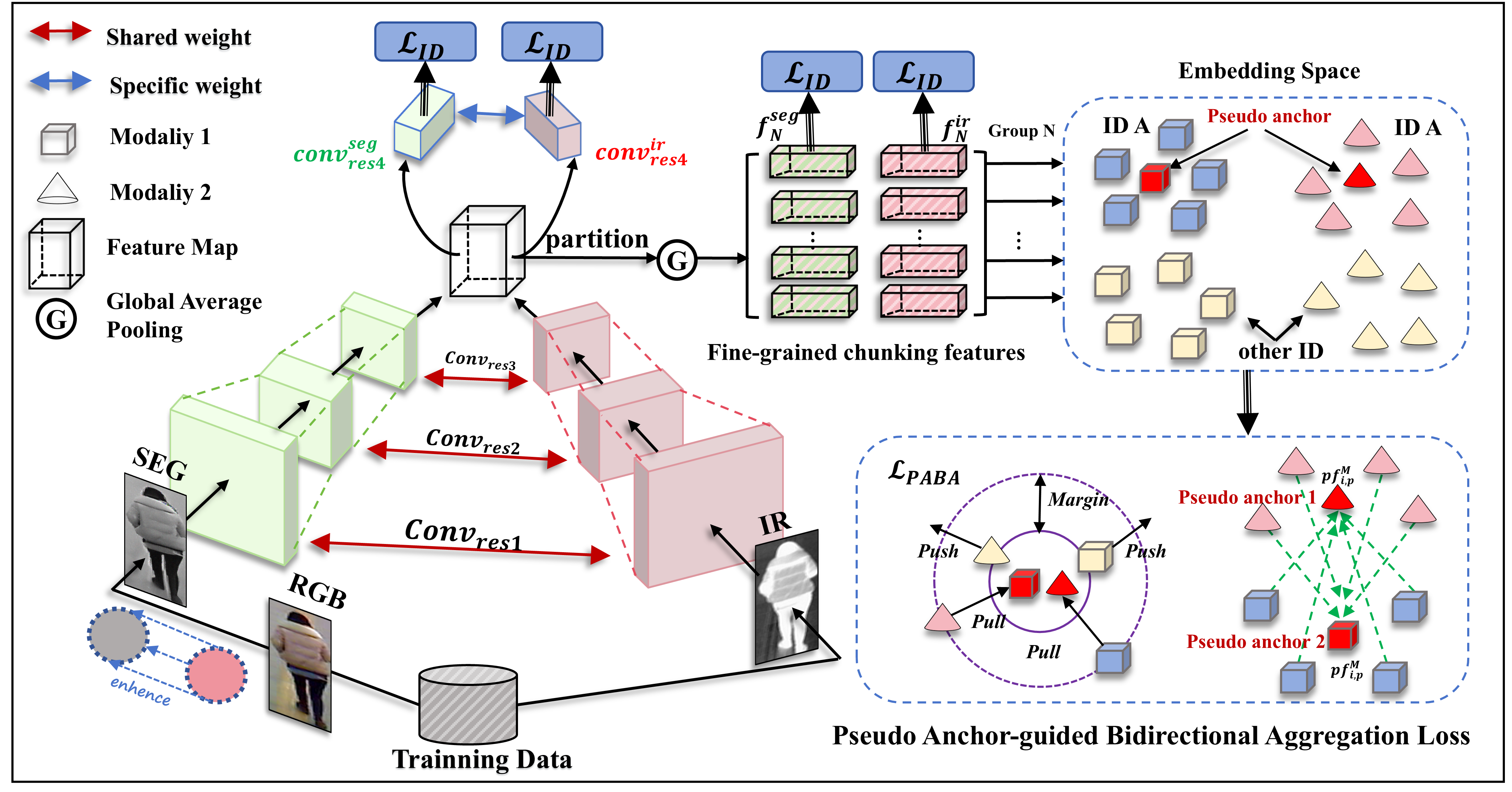}}
\end{minipage}
\caption{The overall architecture of the proposed SEPG-Net. \textbf{First}, we reduce the spectral discrepancies between the RGB and the IR images with channel transformation as well as frequency-domain enhancement, so SEG images are generated. \textbf{Then}, we utilise a dual-stream weight-shareable network to extract features. 
\textbf{Finally}, identity loss and PABA loss are employed to supervise the whole learning process.}
\label{fig:2}
\end{figure*}

To cope aforementioned issues, we propose a novel \textbf{S}pectral \textbf{E}nhancement and \textbf{P}seudo-anchor \textbf{G}uidance \textbf{Net}work \textbf{(SEPG-Net)} for the VI-ReID task. Unlike GAN-based approaches, we first transform RGB/visible features to a more homogeneous greyscale space. Following this, we extract the high-level semantic information from the frequency domain of the RGB images to enhance contour representations. Finally, \textbf{S}emantically \textbf{E}nhanced \textbf{G}rey \textbf{(SEG)} images are generated, which not only approximate the infrared modality but also maintain the original structural information of RGB images. The proposed method enhances cross-spectral features through a simple yet efficient design, which proves to be more conducive to network training against GAN-based methods. In addition, we improve the existing centre aggregation loss by introducing a unique \textbf{P}seudo \textbf{A}nchor-guided \textbf{B}idirectional \textbf{A}ggregation loss (\textbf{PABA}). The PABA loss learns cross-modality features at fine-grained level, using specific pseudo-anchor of each modality to attract hetero-modality features of same instances and push away the features of different instances. The use of PABA enable the network to capture more fine-grained modality-consistent representations while emphasising discriminative identity embeddings.
The main contributions of this paper are summarised as follows:
\begin{itemize}
\item We propose a simple yet efficient strategy for spectral enhancement. For the first time, frequency-domain and greyscale information are used simultaneously to reduce cross-modality discrepancies.
\item We introduce a novel cross-modality constraint, PABA, which explores fine-grained coherence representations while retaining intrinsic discriminative information.
\item Experimental results on the SYSU-MM01 \cite{b21} and RegDB \cite{b20} datasets demonstrate that our method outperforms other state-of-the-art methods.

\end{itemize}

\section{Method}
In this section, we first provide a detailed description of the spectral enhancement process. Then, we introduce the dual-stream structure for modality features extraction. Finally, we explain the working mechanism of the PABA loss. Fig. \ref{fig:2} demonstrates the overall architecture of our proposed method.

\subsection{Semantically Enhanced Grey Images Generation}

Infrared images contain more silhouette detail, while visible images contain more color information \cite{b14, b15, b31, b32, b33, b34, b35, b36,  b37}. As shown in Fig. \ref{fig:3}, we introduce an end-to-end spectral feature enhancement strategy aiming at efficiently generating images that approximate the infrared modality from visible images. This strategy reduces the large spectral gap while preserving complete pedestrian features. Concretely, given a visible image $Y_{vis}$ containing $R$, $G$, and $B$ channels, we introduce a transformation function $\tau(\cdot)$ to traverse each channel and generate the greyscale image $Y_{grey}$, and then replicate it to three channels (get $Y_{grey3}$) in order to match the input infrared images. The above process can be formulated as:
\begin{equation}
Y_{vis} (R,G,B)\overset{\tau (\cdot )}{\rightarrow} Y_{grey}\overset{copy}{\rightarrow} Y_{grey3},
\end{equation}
\begin{equation}
\tau (x )=\alpha \cdot  R(x)+\beta \cdot G(x)+\gamma \cdot B(x),
\end{equation}
where the values of $(\alpha,\beta,\gamma)$ tuple are $(0.299, 0.587, 0.114)$. In addition, several studies have demonstrated that the phase component of Fourier transform contains high-level semantic information, such as body silhouettes \cite{b16, b17, b18}. Inspired by this, we extract the phase information from the frequency domain to enhance the silhouette representation. The generation of semantically enhanced grey image $Y_{seg}$ can be represented as:

\begin{figure*}[t]
\begin{minipage}[b]{1.0\linewidth}
  \centering
  \centerline{\includegraphics[width=0.7\columnwidth]{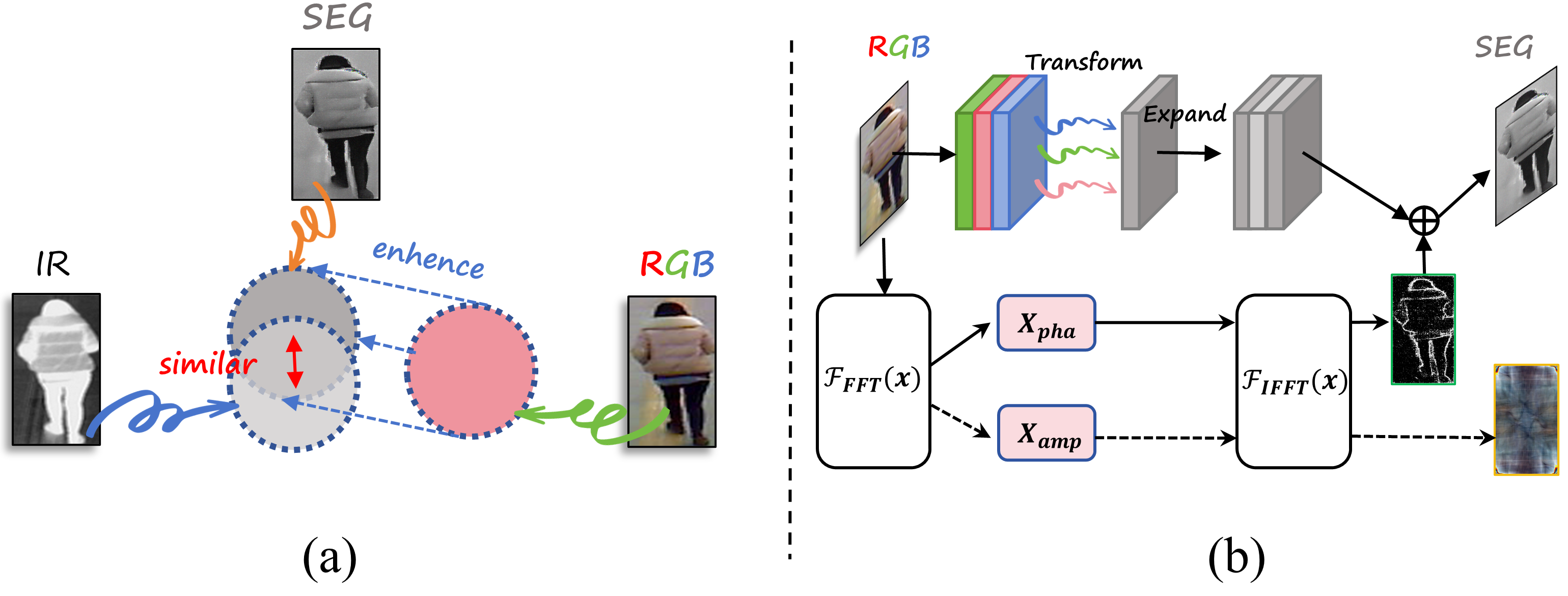}}
\end{minipage}
\caption{(a) We perform spectral feature enhancement by converting visible/RGB image into SEG image, and the enhanced image is similar in style to infrared (IR) image. (b) Details of SEG image generation.}
\label{fig:3}
\end{figure*}

\begin{equation}
X_{pha},X_{amp} =\mathcal{F}_{\mathrm{FFT}}(Y_{vis} ),
\end{equation}
\begin{equation}
Y_{seg} =\mathcal{F}_{\mathrm{IFFT}}(X_{pha})+Y_{grey3},
\end{equation}
where $\mathcal{F}_{\mathrm{FFT}}(\cdot)$ and $\mathcal{F}_{\mathrm{IFFT}}(\cdot)$ denote the Fourier transform and its inverse operation, respectively, and $X_{pha}$ and $X_{amp}$ represent the phase and amplitude components.

\subsection{Weight-Shareable Dual-Stream Network}
To alleviate the remaining modality discrepancies between SEG and IR images, we employ a weight-shareable dual-stream network to extract shared modality representations while preserve the specific modality cues. Specifically, the infrared image input is denoted as $Y_{ir}\in \mathbb{R} ^{H\times W\times C}$ ($H$, $W$, and $C$ denote height, width, and channel, respectively), we apply the first three layers of ResNet-50 \cite{b19} with shared weights to extract the shared features of $Y_{seg}$ and $Y_{ir}$. This process can be represented as the following arithmetic operations:
\begin{equation}
F_{shared}= Conv_{res}^{3} (Conv_{res}^{2}(Conv_{res}^{1}([Y_{ir},Y_{seg} ]_{o}^{b} ))),
\end{equation}
\begin{equation}
F_{shared}^{seg},F_{shared}^{ir}= [F_{shared}]_{h}^{b},
\end{equation}
where $[\cdot]_{o}^{b}$ and $[\cdot]_{h}^{b}$ represent the c\textbf{O}ncatenation and c\textbf{H}unking operations along the \textbf{b}atch dimension. In addition, we duplicate the last layer of ResNet-50 to extract higher-level modality-specific cues $F_{specific }^{seg}$
and $F_{specific }^{ir}$. Unlike previous methods \cite{b5, b6, b7, b8, b9, b10, b11, b12, b13, b38}, the designed structure introduce only one additional layer for cross-modality feature extraction, which avoids extensive computational overhead.

\subsection{Pseudo Anchor-Guided Bidirectional Aggregation Loss}

To further mine modality-consistent representations while maintaining the discrimination of inter-class features, we propose a Pseudo Anchor-guided Bidirectional Aggregation (PABA) loss. It sets a pseudo-anchor for each modality to attract samples from the opposite modality. Compared to the standard triplet loss \cite{b30}, the PABA loss can effectively reduce the intra-class cross-modality discrepancies. Specifically, we split $F_{shared}^{seg}$ and $F_{shared}^{ir}$ into $N$ horizontal parts and obtain chunking features $F_{shared}^{seg}=[f_{1}^{seg},f_{2}^{seg},\cdots, f_{N}^{seg}  ]$ and $F_{shared}^{ir}=[f_{1}^{ir},f_{2}^{ir},\cdots ,f_{N}^{ir}]$. The PABA loss can be formulated as follows:

\begin{equation}
\begin{split}
& L_{PABA}^{(Ma,Mb)}(i)=\frac{1}{K} \sum_{p=1}^{K} max([max D(pf_{i,p}^{Ma},f_{i,p}^{Mb}  )- \\
& \qquad \qquad \qquad min_{p\ne q}D(pf_{i,p}^{Ma},f_{i,q}^{Mb}  ) +margin],0),
\end{split}
\end{equation}
\begin{equation}
L_{PABA}^{\rightleftharpoons }   =\sum_{i=1}^{N} (L_{PABA}^{(seg,ir)} (i)+L_{PABA}^{(ir,seg)} (i)),
\end{equation}
where $Ma$ and $Mb$ represent different modality categories and $K$ is the number of identities. $pf_{i,p}$ denotes the pseudo-anchor from $i$-th part, which is the arithmetic centre of features with same identity and modality in a mini batch. 

We achieve cross-modality bidirectional aggregation within each set of chunking features. Compared to the cross-centre loss in \cite{b9, b10, b11}, this fine-grained design leads to a more compact modality-consistent representations. In addition, we adopt cross-entropy loss as the identity (ID) loss to supervise the modality-specific features as well as the chunking features, and the total loss can be denoted as:
\begin{equation}
L_{ALL}=\lambda _{1}\cdot  L_{ID}^{specific}+\lambda _{2}\cdot L_{PABA}^{\rightleftharpoons }+\lambda _{3}\cdot\sum_{i=1}^{N} L_{ID}^{chunks}(i)
\end{equation}
where $\lambda _{1}$, $\lambda _{2}$, and $\lambda _{3}$ are hyper-parameters to balance the weight of each loss function.

\section{Experiments}

\subsection{Datasets, Evaluation Protocols, and Implementation}\label{AA}
\textbf{Datasets.} The performance of SEPG-Net is evaluated on two public benchmark VI-ReID datasets: RegDB \cite{b20} and SYSU-MM01 \cite{b21}. The RegDB dataset contains 402 identities, with each identity represented by 10 visible and 10 infrared images. RegDB has two test modes: V-2-I (visible-to-infrared) and I-2-V (infrared-to-visible). The SYSU-MM01 dataset is collected with 6 cameras (3 from indoor and 3 from outdoor), comprising a total of 302,420 visible and infrared images from 491 identities. The test modes of SYSU-MM01 are categorized as All-search and Indoor-search.

\textbf{Evaluation Protocols.} Following the standard practices in the ReID community \cite{b1, b2, b3, b4, b5}, we utilize Rank-1 (Probability that the query target appears first in the query results) and mAP (mean Average Precision) to comprehensively evaluate the performance of SEPG-Net and other baseline methods.

\textbf{Implementation Details.} Our SEPG-Net employs a ResNet-50 \cite{b18} as the backbone, pre-trained on ImageNet \cite{b22}. The network is deployed on the Nvidia RTX 4090 GPU for training and testing. In the training phase, each minibatch contains 8 identities, and each identity represented by 7 visible and 7 infrared images. We set the initial learning rate to 0.03, and it decays at the 30th and 70th epoch by a factor of 0.1 (120 epochs in total). The number of horizontal chunks $\textit{N}$ is set to 12. The values of the parameters \textit{margin}, $\lambda _{1}$, $\lambda _{2}$, and $\lambda _{3}$ are set to 0.5, 1, 2.5, and 1, respectively.

\begin{table}[t]
\caption{The Comparison Results of SEPG-Net and State-of-The-Art methods on RegDB and SYSU-MM01 Datasets.}
\centering
\resizebox{\linewidth}{!}{
\begin{tabular}{c|c|cccc|cccc}
\hline
\hline
\multirow{3}{*}{Method} & \multirow{3}{*}{Venue} & \multicolumn{4}{c|}{RegDB}                         & \multicolumn{4}{c}{SYSU-MM01}                      \\ \cline{3-10} 
 &
   &
  \multicolumn{2}{c|}{V-2-I} &
  \multicolumn{2}{c|}{I-2-V} &
  \multicolumn{2}{c|}{All-search} &
  \multicolumn{2}{c}{Indoor-search} \\ \cline{3-10} 
                        &                       & Rank-1 & \multicolumn{1}{c|}{mAP}  & Rank-1 & mAP  & Rank-1 & \multicolumn{1}{c|}{mAP}  & Rank-1 & mAP  \\ \hline
AlignGAN \cite{b5}           & ICCV 19               & 57.9   & \multicolumn{1}{c|}{53.6} & 56.3   & 53.4 & 42.4   & \multicolumn{1}{c|}{40.7} & 45.9   & 54.3 \\
X-Modal \cite{b6}            & AAAI 20               & 62.2   & \multicolumn{1}{c|}{60.2} & -      & -    & 49.9   & \multicolumn{1}{c|}{50.7} & -      & -    \\
NFS \cite{b23}                 & CVPR 21               & 80.5   & \multicolumn{1}{c|}{72.1} & 77.9   & 69.8 & 56.9   & \multicolumn{1}{c|}{55.5} & 62.7   & 69.8 \\
FMCNet \cite{b24}              & CVPR 22               & 89.1   & \multicolumn{1}{c|}{84.4} & 88.4   & 83.9 & 66.3   & \multicolumn{1}{c|}{62.5} & 68.2   & 74.1 \\
CMIT \cite{b12}               & TMM 23                & 88.8   & \multicolumn{1}{c|}{\textcolor{blue}{\textbf{88.5}}} & 84.6   & 83.6 & 70.9   & \multicolumn{1}{c|}{65.5} & 73.3   & 77.2 \\
SFANet \cite{b25}              & TNNLS23               & 76.3   & \multicolumn{1}{c|}{68.0}   & 70.2   & 63.8 & 65.7   & \multicolumn{1}{c|}{60.8} & 71.6   & 80.1 \\
PMWGCN \cite{b26}              & TIFS 24               & 90.6   & \multicolumn{1}{c|}{84.5} & 88.8   & 81.6 & 66.8   & \multicolumn{1}{c|}{64.9} & 72.6   & 76.2 \\
CAJ \cite{b27}                & TPAMI 24              & 85.7   & \multicolumn{1}{c|}{79.7} & 84.9   & 77.8 & 71.5   & \multicolumn{1}{c|}{68.2} & 78.4   & \textcolor{blue}{\textbf{82.0}}   \\
DARD \cite{b28}               & TIFS 24               & 86.2   & \multicolumn{1}{c|}{85.4} & 85.5   & \textcolor{blue}{\textbf{85.1}} & 69.3   & \multicolumn{1}{c|}{65.7} & \textcolor{blue}{\textbf{79.6}}   & 60.3 \\
DCPLNet \cite{b13}            & TII 24                & \textcolor{blue}{\textbf{94.2}}   & \multicolumn{1}{c|}{87.3} & \textcolor{blue}{\textbf{91.7}}   & 84.8 & \textcolor{blue}{\textbf{74.0}}     & \multicolumn{1}{c|}{\textcolor{blue}{\textbf{70.4}}} & 78.3   & 81.9 \\
\textbf{Ours*  }                 & -                     & \textcolor{red}{\textbf{97.7}}   & \multicolumn{1}{c|}{\textcolor{red}{\textbf{90.8}}} & \textcolor{red}{\textbf{95.8}}   & \textcolor{red}{\textbf{90.4}} & \textcolor{red}{\textbf{76.4}}   & \multicolumn{1}{c|}{\textcolor{red}{\textbf{71.3}}} & \textcolor{red}{\textbf{84.5}}   & \textcolor{red}{\textbf{86.2}}\\ \hline \hline
\multicolumn{8}{l}{$^{\mathrm{a}}$ \textcolor{red}{Red} and \textcolor{blue}{blue} represent the best and second-best results, respectively.}
\end{tabular}}
\label{tab:1}
\end{table}

\subsection{Comparison With State-of-the-Art Methods}
To validate the effectiveness of SPEG-Net, several state-of-the-art methods are introduced for comparison \cite{b5,b6,b23,b24,b12,b25,b26,b27,b28,b13}. Table \ref{tab:1} reports the optimal performance achieved by the proposed  SPEG-Net, which surpasses the second-best method on the RegDB dataset by 3.5\% and 4.1\% in Rank-1 and by 2.3\% and 5.3\% for mAP in V-2-I and I-2-V modes, respectively. Moreover, SEPG-Net achieves the best results on the SYSU-MM01 dataset.  This performance can be attributed to two key factors: i) The transformation from RGB to SEG effectively reduce the modality discrepancies at the data level; ii) More fine-grained consistent embeddings are explored by guiding local cross-modality alignment through pseudo anchors. These experimental results strongly demonstrate the superiority of SEPG-Net.

\begin{table}[t]
\centering
\caption{Ablation Experiments of SEPG-Net on SYSU-MM01 Dataset (All-Search Mode).}
\begin{tabular}{cccc|cc}
\hline \hline
baseline & +SE & +CC & +PABA & Rank-1 & mAP  \\ \hline
\checkmark        &     &     &       & 69.6   & 66.8 \\
\checkmark        & \checkmark   &     &       & 72.3   & 69.4 \\
\checkmark        & \checkmark   & \checkmark   &       & \textcolor{blue}{\textbf{73.8}}   & \textcolor{blue}{\textbf{69.7}} \\
\checkmark        & \checkmark   &     & \checkmark     & \textcolor{red}{\textbf{76.4}}   & \textcolor{red}{\textbf{71.3}} \\ \hline \hline
\label{tab:2}
\end{tabular}
\end{table}

\subsection{Ablation Study}
\begin{figure}[]
\begin{minipage}[b]{1.0\linewidth}
  \centering
  \centerline{\includegraphics[width=\columnwidth]{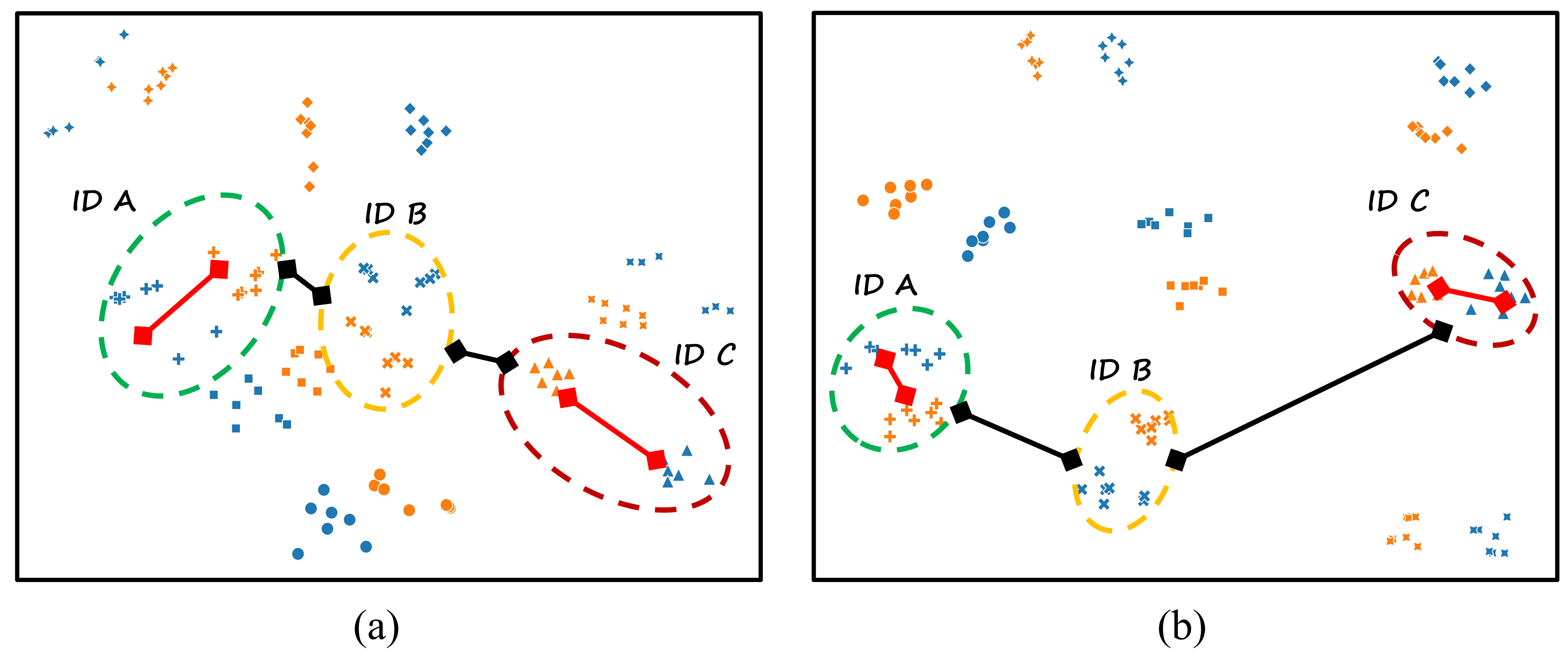}}
\end{minipage}
\caption{t-SNE feature visualisation of baseline (a) and SEPG-Net (b) on SYSU-MM01. Blue and orange represent the infrared and visible modalities, respectively.}
\label{fig:4}
\end{figure}

We also conduct ablation experiments, as shown in Table \ref{tab:2}. The experimental setups are as follows: the baseline includes only the dual-stream backbone supervised with the ID loss. The terms of +SE, +CC and +PAPB represent the application of spectral enhancement, cross-centre loss \cite{b9} and PABA loss, respectively. As observed, the baseline achieves 69.6\% Rank-1 and 66.8\% mAP on the SYSU-MM01 dataset (All-search mode). When the proposed spectral enhancement strategy is utilized to reduce the cross-modality gap, the performance improves by 2.7\% Rank-1 and 2.6\% mAP. Moreover, we compare the classical cross-centre loss with the proposed PABA loss, showing that the PABA leads to an additional 2.6\% improvement in Rank-1 and a 1.6\% gain in mAP compared to the cross-centre loss. Furthermore, the t-SNE \cite{b29} feature visualisation results in Fig. \ref{fig:4} demonstrate that SEPG-Net encourages more compact modality-consistent representations while increasing inter-class discrimination.

\section{Conclusion}
SEPG-Net, a simple yet effective architecture, is developed for VI-ReID task. First, we observe and analyse the spectral discrepancies between infrared and visible images and propose an end-to-end spectral enhancement strategy which uses frequency domain information and greyscale space to bridge the gap from RGB images to infrared images. Furthermore, we rethink the cross-modality learning process of previous VI-ReID approaches and propose a PABA loss paradigm, which endeavours to promote cross-modality bidirectional aggregation at a fine-grained level while preserving discriminative identity information. We compare SEPG-Net's performance with other well-known methods on the RegDB and SYSU-MM01 datasets, and conduct ablation experiments on our key designs. The experimental results demonstrate that SEPG-Net outperforms state-of-the-art methods.

\section{Acknowledgement}
This work was supported by National Nature Science Foundation of China (grant no. U21A6003). (\textsuperscript{\dag}Zhihao Chen and Yiyuan Ge  contribute equally to this work) (*Corresponding author: Ziyang Wang.)

\end{document}